\documentclass[10pt,twocolumn,letterpaper]{article}

\usepackage[pagenumbers]{cvpr} 
\usepackage{caption}
\usepackage{subcaption}
\usepackage{tabularx}
\usepackage[dvipsnames]{xcolor}

\definecolor{cvprblue}{rgb}{0.21,0.49,0.74}
\usepackage[pagebackref,breaklinks,colorlinks,citecolor=cvprblue]{hyperref}
\usepackage{caption} 

\newcommand{\insertfig}{\includegraphics[width=\linewidth]{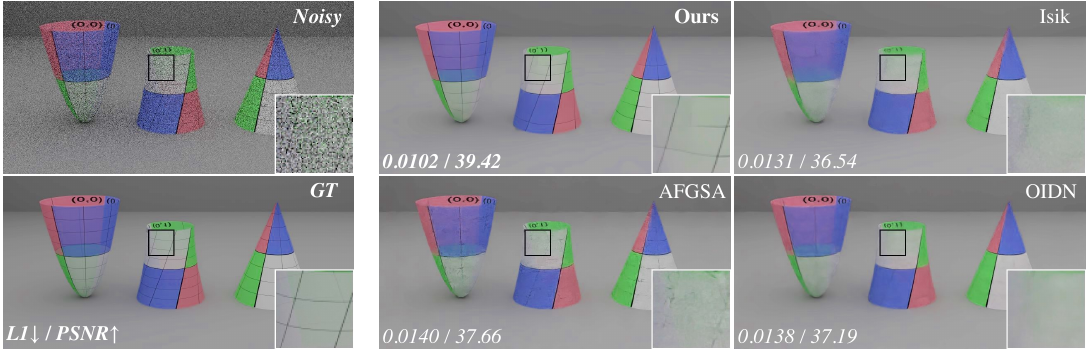}\captionof*{figure}{We present a denoiser based on a pixel-space diffusion model. Because our method has a strong prior of what a real image looks like, it can generalize better on out of distribution images. In the example above, notice how competing methods produce unwanted artifacts like splotchy/blurry regions and missing/broken-up lines on the surface texture. Our method will consistently produce an image that looks like a real image (e.g. minimal unwanted artifacts), while adhering to the conditioning buffers. Results shown on a 4spp test image.}}

\title{Denoising Monte Carlo Renders with Diffusion Models \\ \bigskip \insertfig}

\author{Vaibhav Vavilala\\
UIUC\\
\and
Rahul Vasanth\\
UIUC\\
\and
David Forsyth\\
UIUC\\
}

\begin{document}
\maketitle

\begin{abstract}
Physically-based renderings contain Monte-Carlo noise, with variance that increases as the number of rays per pixel decreases. This noise, while zero-mean for good modern renderers, can have heavy tails (most notably, for scenes containing specular or refractive objects). Learned methods for restoring low fidelity renders are highly developed, because suppressing render noise means one can save compute and use fast renders with few rays per pixel. We demonstrate that a diffusion model can denoise low fidelity renders successfully. Furthermore, our method can be conditioned on a variety of natural render information, and this conditioning helps performance. Quantitative experiments show that our method is competitive with SOTA across a range of sampling rates. Qualitative examination of the reconstructions suggests that the image prior applied by a diffusion method strongly favors reconstructions that are ``like'' real images -- so have straight shadow boundaries, curved specularities and no ``fireflies.'' 
\end{abstract}

\section{Introduction}\label{sec:introduction}
The image produced by a physically-based renderer is the value of a random variable. Typically, the mean of this random variable is the (unknown) true result; but light transport effects mean that the variance -- the result of using ``too few rays'' -- is complicated and heavy tailed. Analogous effects appear in very low light photography. Increasing $N$ -- the number of rays per pixel -- quickly results in diminishing returns, because the variance goes down as $1/N$, and so there is a significant literature that aims to suppress render noise. This paper shows that a method based on diffusion is quantitatively competitive with the state of the art (SOTA) while producing images that differ strongly in qualitative aspects.

Obvious solutions to render noise fail. Real-world scenes in
the film industry are too large to fit into GPU RAM and must be
rendered on a CPU. Often even hundreds to thousands of rays per
pixel may fail to reach an artist's desired level of quality.
Problems are caused by the presence of desirable but complex light transport phenomena such as indirect specular, large numbers of light sources, subsurface scattering, and volumetric effects. 

The concept of relying on pretrained foundation models has been extensively explored for image restoration~\cite{saharia2022palette,li2023diffusion}. Diffusion models have successfully removed JPEG noise and film grain; images can be recolored; and local objects can be inpainted. Even the success of large language models relies on a foundation model trained for next token prediction, then finetuned for particular tasks (for example, chatbots like ChatGPT relying on GPT-4 as the backbone). \textbf{Until now, large foundation models have not been applied to denoising Monte Carlo renders}. 

There are multiple reasons for this. For one, the area has matured - high quality denoisers already exist, some geared towards high quality and others towards speed. Another reason is that there are known workarounds when existing denoisers create problems. Artists can sample scenes for longer (requiring additional compute and slowing artist iteration) or manually touch-up unwanted artifacts in finishing software like Photoshop or Nuke. 

Further, it is not obvious that image foundation models can handle high dynamic range gracefully, since they are trained on images with per-pixel radiance from $[0-1]$. Ray traced images in contrast store the per-pixel colors in linear space, which may exceed $1$ if rendering a bright or shiny object. Subsequent processing such as gamma correction, tonemapping, compositing, and color grading bring the final values to $[0-1]$, suitable for standard dynamic range (SDR) display. Denoisers must process inputs in a manner respecting high dynamic range to be useful in practice. \textbf{Across the billions of images foundation models are trained on, a variety of post-processing will be reflected} in the datasets such that the result sits in the range $[0-1]$. We believe that \textbf{this wide array of tonemapping, geometry, textures, and lighting is an asset if the foundation model can be utilized effectively.} As we show, images denoised with foundation models can often look better than existing methods.


Another potential pitfall is that large scale training sets will include images with unwanted lossy compression artifacts, film grain, image rescaling, motion blur, and other phenomena associated with the capture and storage of imagery. It's not obvious whether an image can be effectively denoised without unwanted artifacts or hallucinations leaking into the result. In practice, we don't notice this to be a problem in our evaluation.

Expensive data collection is another concern - it is extremely compute-intensive to generate 3D scenes and render them with a path tracer to an acceptable level of quality. A single image may require tens to hundreds of CPU hours to converge, and tens of thousands of frames are required to build a sufficient training set. While high quality rendering software exists (such as RenderMan used in VFX houses, PBRT and MITSUBA for research~\cite{christensen2018renderman,pharr2023physically,nimier2019mitsuba}), high quality datasets are not freely available. The research community would benefit from large open source renders of practical scenes with feature buffers at low sample counts, and accompanying high sample target renders. 

A problem we aim to address in this work is that existing denoisers frequently struggle to generalize to scenes with phenomena different from the training set (for example different hair colors/styles, volumetric effects, and novel textures). That may result in users having to collect more training data to retrain an existing denoiser (a computationally demanding effort). Alternatively, denoiser limitations may result in additional artist effort to detect and fix problems manually - or simply accepting a lower-quality product. 

Advancing the SOTA for denoising may alleviate some of these problems. This paper is firmly in the high quality regime (as opposed to focusing on speed), as we are utilizing a multi-pass diffusion model and evaluating against the best existing methods, which are all 1-pass neural networks. We however present suggestions for making our model run much faster. 
\\

\begin{figure*}
  \centering
  \includegraphics[width=16cm]{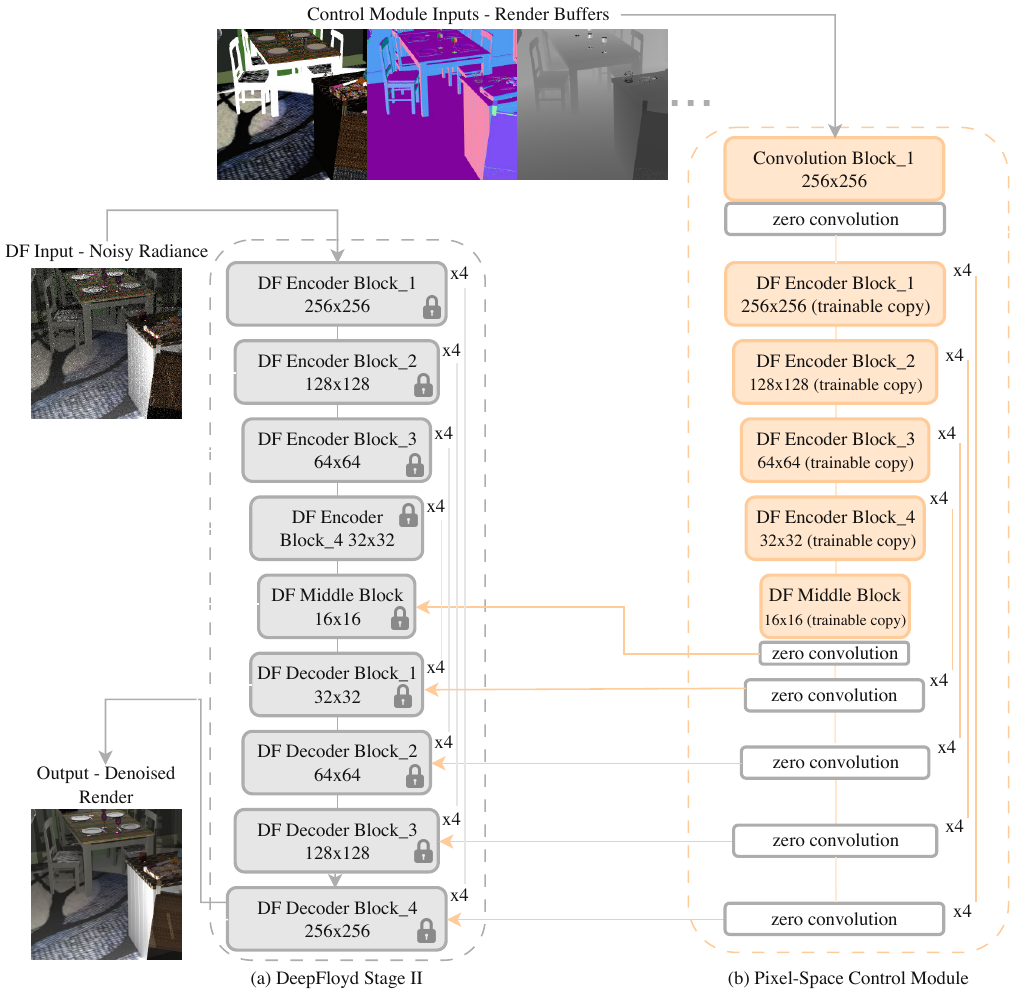}
  \caption{\textbf{Overview of our method.} We leverage a pretrained pixel-space diffusion model, DeepFloyd Stage II~\cite{DeepFloydIF}, as our base synthesizer that is fixed during training. It accepts the noisy radiance as well as a forward-diffused copy of the noisy radiance (see methodology section~\ref{sec:method} for details). We introduce a trainable Control Module, analogous to ControlNet~\cite{zhang2023adding}, initialized with the encoder and middle blocks. It accepts all the auxiliary feature buffers from the renderer like albedo, normals, and depth, in addition to the noisy radiance. The outputs of the Control Module are added to the DF decoder blocks at varying spatial resolutions. We utilize zero convolutions to ease early stages of training. Time and prompt encoding not shown for brevity. Our prompt is the empty string during training and inference.}
  \label{fig:cn}
\end{figure*}

Our contributions are:
\begin{enumerate}
  \item We are the first to use large-scale image generation foundation models to denoise MC renders, and we demonstrate that conditioning on render buffers provides essential information to the diffusion model. To do so, we apply the ControlNet architecture to a pixel-based (rather than latent variable based) diffusion model.
  \item Quantitative and qualitative evidence suggest our method is generally better than existing SOTA methods. 
\end{enumerate}

\section{Related Work}\label{sec:related}

Early denoisers relied on linear regression models and hand-designed filters that run quickly. Zwicker et. al. provide an excellent survey on classical approaches \cite{zwicker2015}, while Huo et. al survey deep learning methods on denoising Monte Carlo renderings \cite{huo2021}. The latest advancements in MC denoising reflect the progression of learning-based approaches. We focus our literature review on recent denoising works and controlled image synthesis.

\subsection{Real-time and Interactive Denoisers}

Chaitanya et. al lower temporal noise in animations and are the first to use a U-net \cite{Chaitanya2017,ronneberger2015unet}. Thomas et. al incorporate a U-net, using a low-precision feature extractor and multiple high-precision filtering stages to perform both supersampling and denoising \cite{Thomas2022TemporallySR}. Fan et. al make improvements to efficiency, building on the hierarchical approach utilized by Thomas et. al by predicting a kernel for a single channel and incorporate temporal accumulation \cite{Fan_2021}. Lin et. al employ a path-based approach \cite{Lin2020PathbasedMC}.

Meng et. al train a network to splat samples onto multi-scale, hierarchical bilateral grids, then denoise by slicing the grid \cite{Meng2020RealtimeMC}. Lee et. al incorporate kernel prediction and temporal accumulation for real time denoising \cite{Lee2021,Lee2024RealTimeMC}. Munkberg et al. build on the approach used by Gharbi et. al, but splat radiance and sample embeddings onto multiple layers. Filter kernels are applied to layers which are composited, improving performance \cite{Munkberg2020}. Munkberg et. al eschew Laplacian pyramids used by Vogels et. al and improve temporal stability compared to Chaitanya et. al. by maintaining a per-sample approach \cite{Munkberg2020, Gharbi2019, Vogels2018}.

Similar to Munkberg et. al, Isik et. al use per-sample information when computing filter weights, though the filters operate on pixel-wise averages. Their network predicts dense features, which are then incorporated in a pairwise-affinity metric that results in per-pixel dilated 2D blur kernels applied iteratively to the low sample radiance. An optional temporal kernel can blend the previous frame's denoised output with the current frame~\cite{Isik2021, Munkberg2020}.

\begin{figure}
  \centering
  \includegraphics[width=8.5cm]{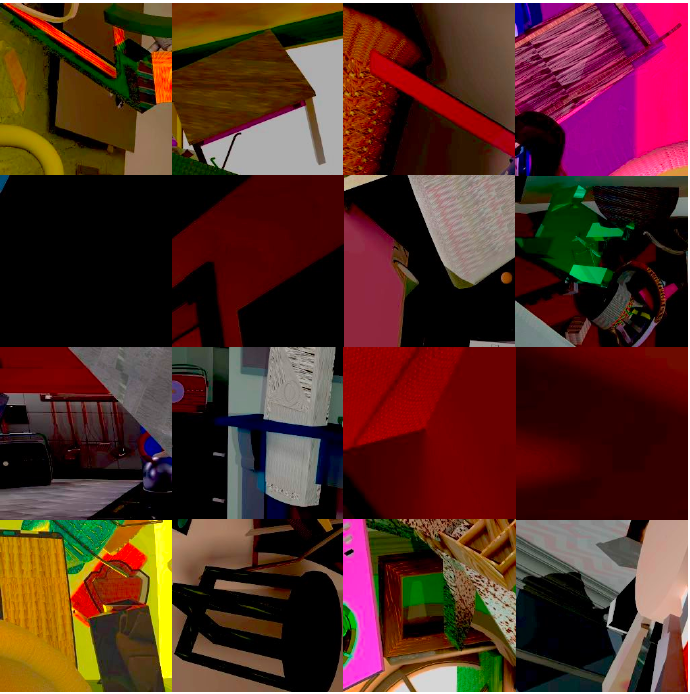}
   \caption{Example images from our procedurally-generated dataset.}

   \label{fig:train_imgs}
\end{figure}

\begin{figure}
  \centering
  \includegraphics[width=8cm]{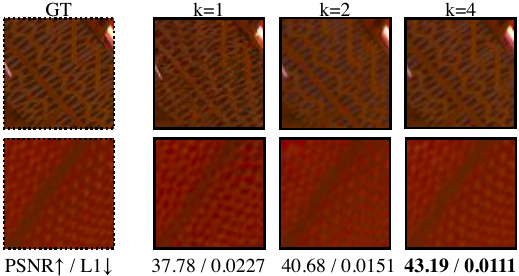}
   \caption{\textit{Best viewed online in color.} Diffusion methods based on a latent variable image representation will not work for render denoising, because the VAE decoding creates significant problems. Existing ControlNet~\cite{zhang2023adding} architectures use a latent representation of the image. The limited size of the VAE dictionary limits the accuracy of very precise pixel space tasks. Upsampling the image helps, but does not remove this effect. In each row, 64x64 cropped training images are shown, highlighting texture. We take the GT training image, upsample by $k$, feed it through a VAE, decode, downsample by $k$ and show the result. We use the default VAE from Stable Diffusion 2.1. Error metrics shown underneath are averaged over 64 random training images. The VAE introduces unacceptable changes to texture. In the first row, the black patches are sharply defined in GT, but blurred after VAE decoding. In the second, the VAE shifts the color of the texture. These changes are partially due to the 8x spatial downsampling factor of VAEs, converting a \textsc{(H,W,3)} image into a \textsc{(H/8, W/8, 4)} dimensional latent code. Thus, existing control mechanisms relying on latent-space diffusion models like ControlNet are not suitable for pixel-space tasks like MC denoising. Even with 4x upscaling, which defeats much of the efficiency gain of the VAE, the PSNR is comparable with existing SOTA denoisers, which effectively caps the quality that can be achieved. Thus, in this work we introduce spatial controls to pixel-space diffusion models.}
\end{figure}

\subsection{Offline Denoisers}

Kalantari et al. utilize a multi-layer perceptron at a per-pixel level to optimize cross-bilateral filter parameters \cite{Kalantari}. 
 
Rather than hand-crafted kernels, Bako et al. use a convolutional neural network to predict filtering kernel weights adaptively at a pixel level, then apply the kernels to the noisy image input. Specular and diffuse components are separately processed. The kernel prediction convolutional network (KPCN) and kernel-prediction approaches are popular as they train quickly and are more robust than predicting colors directly \cite{Bako17}.

Vogels et al. utilize hierarchical pyramid kernels, approximating the behavior of large kernels with a small kernel multi-resolution cascaded filtering strategy. This work extends KPCN to animated sequences, incorporating an adaptive sampling approach with temporal aggregation \cite{Vogels2018}. Instead of inferring weights per pixel, Gharbi et al. use radiance samples, demonstrating their utility for denoising by predicting splatting kernels for each sample \cite{Gharbi2019}. Balint et al. further develop the pyramidal filtering approach with improvements to upsampling, weight predictor networks, and learnable partitioning \cite{Balint2023}. 

Xu et. al show that generative adversarial networks can be used for denoising, eliminating in-between layers used in Bako et. al, but similarly train separate networks to process specular and diffuse components \cite{Xu2019, Bako17}. Yu et. al develop this adversarial approach by processing all components together and incorporate a modified self-attention: auxiliary feature guided self-attention (AFGSA) \cite{Yu2021}. This work effectively implements a global blur kernel since each pixel attends to the features of all pixels via the cross-attention mechanism. Back et. al apply a post-processing network incorporating a self-supervised loss to KPCN, AFGSA, and Xu et. al~\cite{Xu2019} to improve denoising quality \cite{Back22}.

Our work deviates from much of the recent kernel-based approaches in that we predict the colors directly.

\subsection{Diffusion Models}
Stable Diffusion is a foundational model that implements latent diffusion \cite{rombach2022highresolution, ho2020denoising, song2020generative}, whereby a VAE compresses the spatial resolution of the input image, the diffusion model is run in latent space, and the resulting latent code is decoded into the output image. The synthesis is commonly conditioned by text input via cross attention. Additional controls can be introduced, most notably ControlNet~\cite{zhang2023adding}. ControlNet has been shown to facilitate control over latent diffusion models by conditioning image synthesis on arbitrary spatial information like edges, depth, and segmentation. 

Imagen is another successful large-scale image synthesis model that operates in pixel space \cite{saharia2022photorealistic}, with image quality comparable to Stable Diffusion. Imagen relies on three diffusion models: the first synthesizes at 64x64 resolution; the second upsamples to 256 resolution; and the third upsamples to 1024. Thus, super resolution replaces the need for a VAE, easing compute requirements. DeepFloyd is an open source implementation of Imagen~\cite{DeepFloydIF}.  

Wang et. al pass pixel-level features to a pretrained Stable Diffusion model for restoration~\cite{wang2023exploiting}. Yang et. al introduce a pixel-aware cross attention module to latent diffusion models for realistic image super-resolution \cite{yang2023pasd}. Similarly, Instruct-Imagen utilizes cross-attention to condition a pixel space diffusion model on spatial layout~\cite{hu2024instruct}. In contrast, we use concatenation and addition to condition our diffusion model, like ControlNet, but in pixel space.

\section{Method Overview}\label{sec:method}
\subsection{Data}
\textbf{Rendering} We generate approx. 5650 random scenes using a procedure similar to~\cite{Isik2021}. Each scene contains randomly arranged ShapeNet objects, textures, and lighting configurations~\cite{chang2015shapenet}. For each scene, we create a random camera move over the course of 8 frames and render with PBRT-V3 at 256 resolution~\cite{pharr2023physically}. We render with the path-tracing integrator and allow several bounces to capture indirect illumination. For each frame, we render the ground truth $GT^{1}_{\text{raw}}$ at 4096 spp, and render 16 additional frames with feature buffers at 4spp (all with different random seeds to avoid noise correlations). Thus we can train our denoiser with spp $\in \{4, 16, 64\}$. Our auxiliary feature buffers $F$ include normals, albedo, depth, direct specular, indirect specular, direct diffuse, indirect diffuse, roughness, emissive, metallic, and transmission. Variance for each channel is also stored.

\textbf{Data sanitization} Unfortunately, our target renders were still a bit noisy at such a high spp due to indirect illumination noise. Noise2Noise \cite{lehtinen2018noise2noise} establishes that neural networks can learn to denoise even when ground truth is noisy if the training objective (often an L1 or L2 loss) attempts to recover the mean of the distribution. This assumption allows several prior methods to succeed even when ground truth is noisy. However, methods relying on a discriminator such as AFGSA~\cite{Yu2021} are not compatible with noisy ground truth because noise will leak into the generated distribution. Thus we adopt the following procedure to post-process the ground truth images. We render an additional GT image $GT^{2}_{\text{raw}}$ at 4096 spp for each scene, but use a different random seed than $GT^{1}_{\text{raw}}$. We then train a network with only SMAPE loss that accepts a GT render and predicts the other one from the pair that we rendered. This data-cleaning network $C_\theta$ is optimized with:

\begin{equation}
{L_{\text{SMAPE}} = \frac{1}{N}\sum_i \text{SMAPE}(C_{\theta}(F_i,GT^1_{i,\text{raw}}),GT^2_{i,\text{raw}})}
\label{eq:dataClean}
\end{equation}

Where the loss is averaged over $i=1,2,...N$ samples in a minibatch and $GT^1_{i,\text{raw}}$ and $GT^2_{i,\text{raw}}$ are interchanged at random during training. SMAPE is defined as:

\begin{equation}
{\text{SMAPE}(A,B) = \frac{||A-B||_1}{||A||_1 + ||B||_1 + \epsilon}}
\label{eq:smape}
\end{equation}

We adopt the architecture of Isik~\cite{Isik2021} for $C_\theta$, conditioning the input on features but excluding any temporal loss terms. The fully-trained data cleaning network is then run on all the ground truth images we rendered (we just pick one out of each pair of raw GT frames), generating clean GT renders:

\begin{equation}
GT_{i,\text{clean}} = C_\theta(F_i,GT^1_{i,\text{raw}}). 
\label{eq:dataInfer}
\end{equation}

All evaluated methods use $GT_{\text{clean}}$ as the target during training, and have access to the same training, validation, and test splits.

\textbf{Range-compression} Large-scale pretrained diffusion models are trained on images with various degradations and tonemappings applied, ultimately with pixel values ranging from 0 to 1. Our dataset however has useful radiance values well beyond that - we clamp the radiance buffers to 6. The practice of range-compressing in log space is common across all MC denoising works we evaluated. In our case, we observe that gamma-correcting tonemappers produce more natural-looking results consistent with datasets diffusion models were trained on. Thus we conduct initial experiments with the following tonemapper: 
\begin{equation}
{\displaystyle V_{\text{out}}=AV_{\text{in}}^{\gamma }}
\label{eq:gamma}
\end{equation}
We set $A=0.47, \gamma=1/2.4$, fully capturing the $0-6$ range. However, early results showed that log tonemapping produces better results for our model:
\begin{equation}
{\displaystyle V_{\text{out}}=A*log(1+V_{\text{in}})}
\label{eq:log}
\end{equation}
We set $A=0.51$. In our evaluation, we show that our diffusion model can effectively denoise images in a LDR space, and we reverse-tonemap the synthesized result back to HDR space for error metric calculation. All figures in this manuscript are tonemapped with gamma (eqn.~\ref{eq:gamma}), as the details appear more clear.  

\textbf{Additional test data} We hold out 3\% of the data for validation and 5\% for testing. For qualitative evaluation, we render a few PBRT-v3 test scenes that portray practical real-world scenarios\footnote{\url{https://www.pbrt.org/scenes-v3}} though we do not apply post-processing to the ground truth as we did for the training images.

For 4 spp models, we randomly sample one low-res stack of buffers out of 16 available; we sample 4 buffers for the 16 spp models; and use them all for the 64 spp models. The aux buffers are averaged over the selections. 

\subsection{Architecture}
We use DeepFloyd~\cite{DeepFloydIF} (which is based on Google's Imagen~\cite{saharia2022photorealistic}) as our base architecture, and build a control mechanism around it in a manner similar to ControlNet~\cite{zhang2023adding}. Our Control Module, shown in Fig.~\ref{fig:cn}, consists of a few convolutional layers that accept arbitrary numbers of channels (39 aux buffers in our case), the DeepFloyd encoder layers, and the DeepFloyd middle layer. The Control Module generates feature maps at resolutions (256, 128,...16) which are summed with the original DeepFloyd encoder outputs and passed to the DeepFloyd decoder. DeepFloyd modules are not trainable, only the Control Module is, consistent with the original ControlNet. Zero convolutions are initialized to ensure the system produces realistic images from the very beginning. We only use the DeepFloyd Stage II module for this work, which is intended to perform super resolution; we show it can perform image denoising here. We use the smaller \textsc{IF-II-M} model for efficiency, which has 450M parameters. Our base module accepts two 3 channel inputs that are concatenated: a time-dependent noisy radiance buffer $q\left(\mathbf{z}_t \mid \mathbf{x}\right)$, which undergoes the diffusion forward noising process, and that same noisy radiance $x$ without forward noising (because the ${aug\_level}$ parameter is fixed to $0$ during training and inference). The Control Module accepts the feature buffers $F_i$. The forward diffusion process is defined as:

\begin{multline}
q\left(\mathbf{z}_t \mid \mathbf{x}\right)=\mathcal{N}\left(\mathbf{z}_t ; \alpha_t \mathbf{x}, \sigma_t^2 \mathbf{I}\right), \\ \quad q\left(\mathbf{z}_t \mid \mathbf{z}_s\right)=\mathcal{N}\left(\mathbf{z}_t ;\left(\alpha_t / \alpha_s\right) \mathbf{z}_s, \sigma_{t \mid s}^2 \mathbf{I}\right)
\label{eq:fwdNoise}
\end{multline}

where $0 \leq s<t \leq 1, \sigma_{t \mid s}^2=\left(1-e^{\lambda_t-\lambda_s}\right) \sigma_t^2$, and $\alpha_t, \sigma_t$ define a differentiable noise schedule whose log signal-to-noise-ratio, i.e., $\lambda_t=\log \left[\alpha_t^2 / \sigma_t^2\right]$, decreases with $t$ until $q\left(\mathbf{z}_1\right) \approx \mathcal{N}(\mathbf{0}, \mathbf{I})$. For generation, the diffusion model is trained to reverse this forward process. We refer the reader to~\cite{saharia2022photorealistic} for additional details. 

\textbf{Losses} We use the usual losses documented in~\cite{saharia2022photorealistic,DeepFloydIF}, i.e., mean squared error and variational lower bound. We let the text prompt be the empty string during training and inference.

\begin{figure}
  \centering
  \includegraphics[width=6cm]{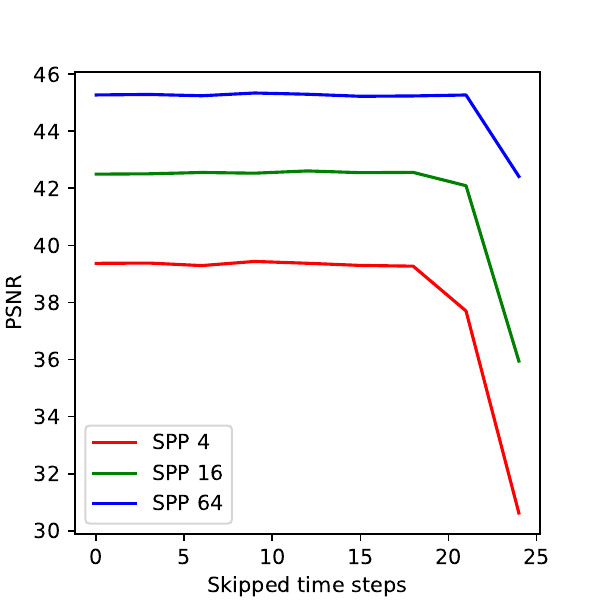}
   \caption{Our method requires approx. 2.8 seconds to denoise a 256x256 image and 63 seconds to denoise an HD 1080x1920 frame (without skipping time steps) on an A40 GPU. We use mixed-precision during inference and \textsc{super27} DDPM sampler in DeepFloyd. This experiment suggests that we can skip around 18 of the 27 denoising time steps with negligible loss in quality, making diffusion models even more practical. At inference time, we can add gaussian noise to the low-spp render via equation~\ref{eq:fwdNoise} instead of starting from pure noise. The gaussian noise strength needs to be sufficient to overcome the variance in the noisy render. Even though diffusion is more expensive than single-pass methods, the cost of denoising remains much smaller than the cost of rendering real-world scenes to convergence, which can be dozens of hours. Even scenes that fit into GPU memory can take dozens of minutes to render to convergence, which dwarfs the cost of denoising low-ray estimates.}

   \label{fig:eval_curves}
\end{figure}

\begin{table*}[h!]
\centering
\caption{\textbf{Quantitatively, our method is competitive with SOTA}. We compare error metrics across different spp settings (4, 16, 64) for various methods on our test set, consisting of 226 sequences of 8 frames each at 256 res. All methods are run in single-frame mode and evaluation metrics do not take into account temporal performance. Our method outcompetes all others in L1, which is computed in HDR $[0-6]$ space, as well as FoVVDP. We are also competitive in other metrics, which evaluate both pixel-wise and perceptual quality. DINO~\cite{oquab2023dinov2, darcet2023vitneedreg} and CLIP measure cosine similarity of the generated and GT global image feature vector. In the \textbf{fourth row}, we disable our Control Module and denoise with an off-the-shelf DeepFloyd Stage II model, presenting the best numbers after ablating the number of skipped denoising time steps and the $aug\_level$ parameter. Applying pure DeepFloyd to low-ray estimates is not successful. \textbf{Fifth row} Thus, conditioning our method with render buffer information via our Control Module makes a significant difference to performance.}
\begin{tabular}{l|ccccccc}
\hline
\textbf{Method} & \textbf{L1 $\downarrow$} & \textbf{PSNR $\uparrow$} & \textbf{LPIPS $\downarrow$} & \textbf{DINO $\uparrow$}~\cite{oquab2023dinov2} & \textbf{CLIP $\uparrow$}~\cite{radford2021learningtransferablevisualmodels} & \textbf{FliP $\downarrow$}~\cite{Andersson2020} & \textbf{FoVVDP $\uparrow$}~\cite{10.1145/3450626.3459831} \\
\hline
AFGSA~\cite{Yu2021} & 0.0279 & 38.672 & 0.1130 & 0.914 & 0.932 & 0.0494 & 8.699 \\
Isik~\cite{Isik2021} & 0.0499 & 38.828 & 0.0871 & 0.945 & 0.955 & \textbf{0.0476} & 8.835 \\
OIDN~\cite{OpenImageDenoise} & 0.0638 & 36.329 & 0.1192 & 0.915 & 0.904 & 0.0691 & 8.645 \\
DeepFloyd-II~\cite{DeepFloydIF} & 0.1009 & 27.583 & 0.3860 & 0.742 & 0.804 & 0.1390 & 6.513 \\
\textbf{Ours} & \textbf{0.0237} & \textbf{39.130} & \textbf{0.0748} & \textbf{0.948} & \textbf{0.965} & 0.0487 & \textbf{8.888} \\
\hline
\end{tabular}
\subcaption*{\textbf{4spp}}
\bigskip
\begin{tabular}{l|ccccccc}
\hline
\textbf{Method} & \textbf{L1 $\downarrow$} & \textbf{PSNR $\uparrow$} & \textbf{LPIPS $\downarrow$} & \textbf{DINO $\uparrow$}~\cite{oquab2023dinov2} & \textbf{CLIP $\uparrow$}~\cite{radford2021learningtransferablevisualmodels} & \textbf{FliP $\downarrow$}~\cite{Andersson2020} & \textbf{FoVVDP $\uparrow$}~\cite{10.1145/3450626.3459831}\\
\hline
AFGSA~\cite{Yu2021} & 0.0184 & 42.076 & 0.0708 & 0.939 & 0.958 & 0.0350 & 9.222 \\
Isik~\cite{Isik2021} & 0.0449 & 42.044 & 0.0560 & 0.964 & 0.972 & 0.0346 & 9.283 \\
OIDN~\cite{OpenImageDenoise} & 0.0537 & 39.869 & 0.0858 & 0.937 & 0.924 & 0.0459 & 9.158 \\
DeepFloyd-II~\cite{DeepFloydIF} & 0.0825 & 28.988 & 0.350 & 0.775 & 0.819 & 0.1076 & 7.150 \\
\textbf{Ours} & \textbf{0.0156} & \textbf{42.343} & \textbf{0.0499} & \textbf{0.965} & \textbf{0.975} & \textbf{0.0338} & \textbf{9.328} \\
\hline
\end{tabular}
\subcaption*{\textbf{16spp}}
\bigskip
\begin{tabular}{l|ccccccc}
\hline
\textbf{Method} & \textbf{L1 $\downarrow$} & \textbf{PSNR $\uparrow$} & \textbf{LPIPS $\downarrow$} & \textbf{DINO $\uparrow$}~\cite{oquab2023dinov2} & \textbf{CLIP $\uparrow$}~\cite{radford2021learningtransferablevisualmodels} & \textbf{FliP $\downarrow$}~\cite{Andersson2020} & \textbf{FoVVDP $\uparrow$}~\cite{10.1145/3450626.3459831} \\
\hline
AFGSA~\cite{Yu2021} & 0.0142 & \textbf{45.091} & 0.0467 & 0.956 & 0.972 & \textbf{0.0260} & 9.562 \\
Isik~\cite{Isik2021} & 0.0433 & 45.055 & \textbf{0.0341} & \textbf{0.977} & \textbf{0.982} & \textbf{0.0260} & 9.593 \\
OIDN~\cite{OpenImageDenoise} & 0.0488 & 42.664 & 0.0640 & 0.954 & 0.940 & 0.0329 & 9.482 \\
DeepFloyd-II~\cite{DeepFloydIF} & 0.0696 & 29.777 & 0.317 & 0.786 & 0.830 & 0.0973 & 7.261 \\
\textbf{Ours} & \textbf{0.0113} & 44.953 & 0.0346 & 0.975 & \textbf{0.982} & 0.0261 & \textbf{9.616} \\
\hline
\end{tabular}
\subcaption*{\textbf{64spp}}
\label{tab:comparison_metrics}
\end{table*}

\begin{figure*}
  \centering
  \includegraphics[width=17cm]{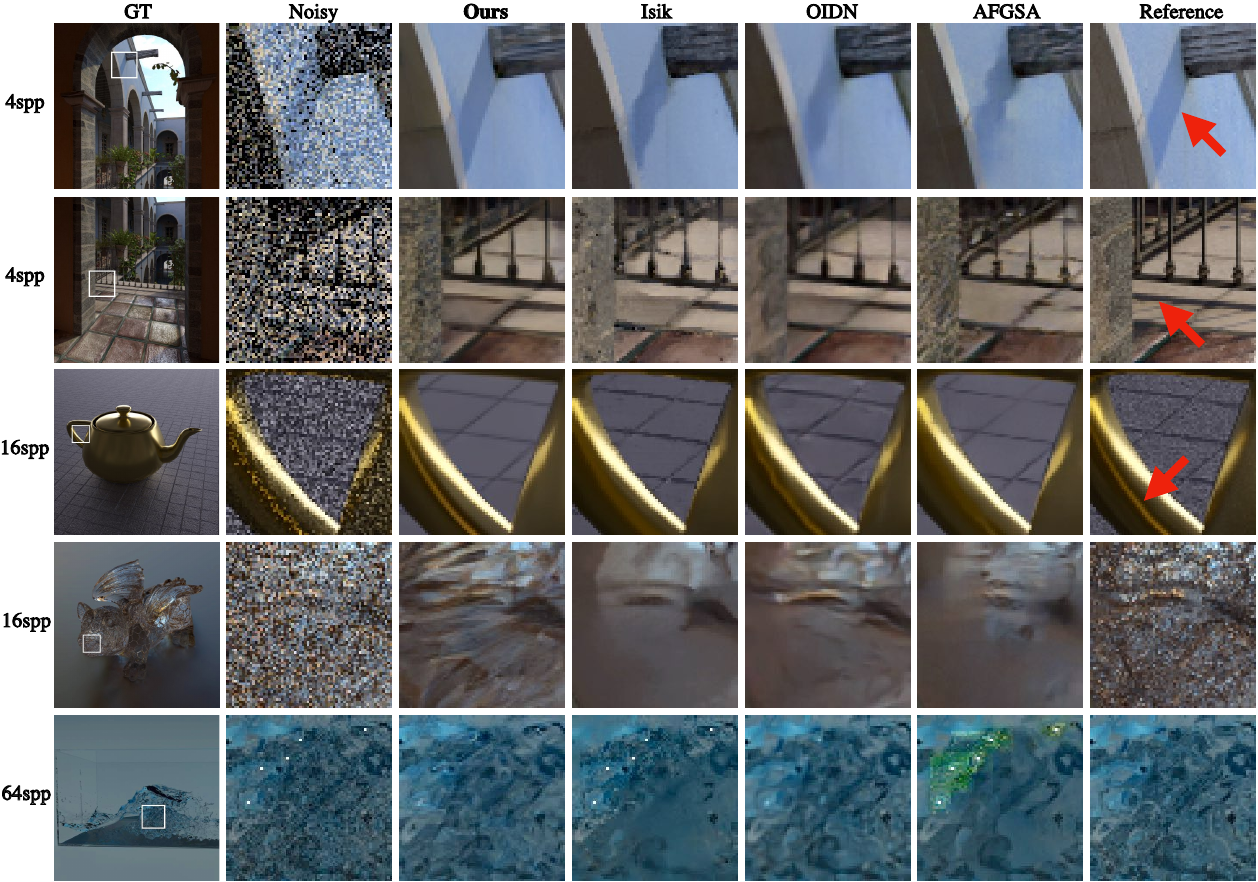}
   \caption{Qualitatively, our reconstructions look like real images, because DeepFloyd has a very strong notion of what an image looks like (it has seen a huge dataset) and because the conditioning buffers offer strong guidelines (e.g. normals, albedo, and depth). In the \textbf{final column}, red arrows point to areas of interest. \textbf{First row.} Notice the straight edge on the shadow (ours; real images tend to have straight shadow edges) compared with blurred or blotchy edges (others). \textbf{Second row.} In undersampled regions, our method fills in the shadow underneath a railing (real images do not have incomplete shadows); other methods render a blurred or incomplete shadow. \textbf{Third row.} Notice the clean sharp highlight on the teacup handle, and smooth highlight boundaries (ours; real images have clean sharp highlight boundaries) compared with absent highlights and blotchy boundaries (others). Notice also aliasing effects in the background, most prominent for OIDN, and absent from our reconstruction. \textbf{Fourth row.} All methods have problems with this specular dragon. Competing methods overblur the dragon's mouth, whereas ours hallucinates plausible details. \textbf{Fifth row.} Fireflies, also known as spike noise, are single very bright pixels, which do not occur in real images. It is rare that fireflies appear at 64 spp in our training set, so AFGSA and Isik fail to remove them. OIDN succeeds in removing them, likely because we use their pretrained model trained on a large dataset.}

   \label{fig:zooms}
\end{figure*}

\textbf{Training \& Inference} We train with AdamW optimizer, mixed precision, and batch size 12. While the model is trained at 256 res, it can run inference at varying spatial resolutions as it relies on convolutional and transformer layers. We use the \textsc{super27} DDPM inference schedule. We experimentally set the ${aug\_level}$ parameter to $0$ during training and inference. This parameter is intended to augment the Stage II input, which typically comes from Stage I. In our case we supply the noisy radiance $\mathbf{x}$ (which is already at the target resolution). We train for 7 days (170 epochs) with initial learning rate $0.00002$ and halved every two days. All experiments are conducted on one NVIDIA A40 GPU. We train an independent network for each sampling rate (4/16/64) though we would expect one appropriately-trained network to succeed on all spp's in practice. We train independent models for competing methods as well. 

\section{Evaluation}
\label{sec:eval}

We focus our comparative evaluation on 3 prior methods known for quality - Isik, AFGSA, and OIDN~\cite{Isik2021,Yu2021,OpenImageDenoise}. Isik and OIDN rely on CNN U-nets, and AFGSA uses a transformer with adversarial loss. We evaluate the best available OIDN pretrained model (\textsc{oidn-version} = \textsc{2.1.0}) and supply it at test time with noisy radiance, albedo, and normals. We retrain Isik and AFGSA with default hyperparameters, all feature buffers, and best-performing model on the held-out validation set used for testing. 

We evaluate each method using two standard metrics: L1 (applied in HDR space) and PSNR (in LDR space). We also evaluate several perceptual metrics: DINO and CLIP feature similarity, LPIPS, FoVVDP, and FliP. For PSNR and perceptual metrics, we tonemap GT and prediction via the following tonemapper, consistent with existing evaluation methods:

\begin{equation}
V_{out} = \frac{V_{in}}{1+V_{in}}
\label{eq:tm3}
\end{equation}

\textbf{Quantitative evaluation} We test each method on 226 sequences of 8 frames each at 256 res in Table~\ref{tab:comparison_metrics}. Across spp $\in \{4, 16,64\}$, our method is the top-performer as measured by L1 and FoVVDP. We are competitive with other methods as measured by PSNR and several perceptual metrics. For our method, we use 27 DDPM steps at mixed-precision. However, in Fig.~\ref{fig:eval_curves}, we ablate a key test-time hyperparameter. We show that we can skip over half the DDPM steps with negligible loss in PSNR, resulting in additional savings. This behavior is consistent across all spp's we tested. 

Finally, in the fourth row of each table in Tab.~\ref{tab:comparison_metrics}, we examine the effects of removing the Control Module and denoising with an off-the-shelf DeepFloyd Stage II model, finding that the results are much worse. We test several skipped time step and $aug\_level$ values, and report the best numbers. We skip 6 time steps and set $aug\_level = 0.5$ in this row. Removing the Control Module hurts quality, thus we conclude it is necessary to obtain good results.

\textbf{Qualitative evaluation} While our error metrics are generally better, the value of pretrained diffusion models becomes clear in our qualitative evaluation. The key finding from our method is that images look more plausible and realistic because the image model has seen billions of images. We demonstrate this in Fig.~\ref{fig:zooms}. In some cases, competing methods produce numbers comparable or slightly better than ours. But upon closer inspection, the details rendered from our diffusion model look more realistic, because our model has a strong prior of what a real image looks like. Thus shadows, specular highlights, and undersampled edges look reasonable. Other methods overblur undersampled regions, fail to remove fireflies even when trained to do so, and hallucinate splotchy patches when pushed outside the training distribution. Our method is not explicitly engineered to remove fireflies or hallucinate nice-looking specularities and shadows; it only knows it should synthesize a realistic image that follows the conditioning.


\section{Conclusion}
\label{sec:Limitations}
Pretrained large-scale image models enable MC denoising that is quantitatively competitive with SOTA and qualitatively more realistic. These models that have seen billions of images are clearly beneficial for the denoising task, where expensive, curated training sets typically number in the tens to hundreds of thousands of scenes. One-step models like GANs and distilled diffusion models may yield significant efficiency gains with acceptable quality trade-offs~\cite{sauer2023stylegan,sauer2024fast}.

One area of future work is what conditioning to use; another is what effects to capture in a training set. Our training set is relatively straightforward, though we test on images with challenging effects. A diverse training set that aggressively oversamples difficult effects might produce better results. Another area of future work is video. As video generators become available~\cite{videoworldsimulators2024,blattmann2023stable}, our approach might usefully be extended to temporally-coherent denoising.

{
    \small
    \bibliographystyle{ieeenat_fullname}
    \bibliography{main}
}
\clearpage
\setcounter{page}{1}
\maketitlesupplementary

\section{Additional Evaluation}
\label{sec:add_eval}
We evaluate our method on the noisebase dataset (\url{https://balint.io/noisebase/datasets/index.html}), which consists of 1024 sequences of 64 frames each, 256-res. Each training and test sample has accompanying feature buffers (depth, normal, and albedo) as well as temporal information like camera parameters and motion vectors that we do not consider here. Each pixel contains per-sample information up to 32spp, which we average to pixel-space at the appropriate sampling rate (as our method operates on pixels and not samples). We then compare our method to other pixel-space methods,~\cite{OpenImageDenoise},~\cite{Yu2021}. The test sets consist of up to 9 scenes with 40-160 full HD frames in each. In this experiment, we train one model to denoise the full spectrum of available sampling rates (1-32spp) for our method and AFGSA~\cite{Yu2021}. We use OIDN's pretrained model as additional evaluation~\cite{OpenImageDenoise}. In our qualitative evaluation, we present several examples showing that our method excels at producing reasonable textures particularly in undersampled regions. Other methods shift the color or hallucinate details very different from the input.  

\begin{table*}[h!]
\centering
\caption{\textbf{Quantitatively, our method is competitive with SOTA}. We compare error metrics across different spp settings (2, 4, 8, 32) for various methods on the noisebase test set, consisting of several scenes at full HD. Across all sampling rates, we are the best method for most metrics, which evaluate both pixel-wise and perceptual quality. DINO~\cite{oquab2023dinov2, darcet2023vitneedreg} and CLIP measure cosine similarity of the generated and GT global image feature vector.}
\begin{tabular}{l|ccccccc}
\hline
\textbf{Method} & \textbf{L1 $\downarrow$} & \textbf{PSNR $\uparrow$} & \textbf{LPIPS $\downarrow$} & \textbf{DINO $\uparrow$}~\cite{oquab2023dinov2} & \textbf{CLIP $\uparrow$}~\cite{radford2021learningtransferablevisualmodels} & \textbf{FliP $\downarrow$}~\cite{Andersson2020} & \textbf{FoVVDP $\uparrow$}~\cite{10.1145/3450626.3459831} \\
\hline
AFGSA~\cite{Yu2021} & 0.163 & 24.8 & 0.436 & 0.946 & 0.887 & 0.167 & 6.34 \\
OIDN~\cite{OpenImageDenoise} & \textbf{0.0990} & \textbf{26.8} & 0.421 & 0.974 & 0.924 & 0.148 & \textbf{7.07} \\
\textbf{Ours} & 0.114 & 26.5 & \textbf{0.401} & \textbf{0.975} & \textbf{0.939} & \textbf{0.147} & 6.94 \\
\hline
\end{tabular}
\subcaption*{\textbf{2spp}}
\bigskip
\begin{tabular}{l|ccccccc}
\hline
\textbf{Method} & \textbf{L1 $\downarrow$} & \textbf{PSNR $\uparrow$} & \textbf{LPIPS $\downarrow$} & \textbf{DINO $\uparrow$}~\cite{oquab2023dinov2} & \textbf{CLIP $\uparrow$}~\cite{radford2021learningtransferablevisualmodels} & \textbf{FliP $\downarrow$}~\cite{Andersson2020} & \textbf{FoVVDP $\uparrow$}~\cite{10.1145/3450626.3459831}\\
\hline
AFGSA~\cite{Yu2021} & 0.0992 & 27.3 & 0.394 & 0.963 & 0.914 & 0.125 & 7.07 \\
OIDN~\cite{OpenImageDenoise} & 0.0769 & 28.2 & 0.392 & 0.980 & 0.936 & 0.123 & \textbf{7.54} \\
\textbf{Ours} & \textbf{0.0748} & \textbf{28.5} & \textbf{0.368} & \textbf{0.982} & \textbf{0.950} & \textbf{0.114} & 7.50 \\
\hline
\end{tabular}
\subcaption*{\textbf{4spp}}

\bigskip
\begin{tabular}{l|ccccccc}
\hline
\textbf{Method} & \textbf{L1 $\downarrow$} & \textbf{PSNR $\uparrow$} & \textbf{LPIPS $\downarrow$} & \textbf{DINO $\uparrow$}~\cite{oquab2023dinov2} & \textbf{CLIP $\uparrow$}~\cite{radford2021learningtransferablevisualmodels} & \textbf{FliP $\downarrow$}~\cite{Andersson2020} & \textbf{FoVVDP $\uparrow$}~\cite{10.1145/3450626.3459831} \\
\hline
AFGSA~\cite{Yu2021} & 0.0834 & 28.8 & 0.361 & 0.975 & 0.938 & 0.104 & 7.61 \\
OIDN~\cite{OpenImageDenoise} & 0.0611 & 29.7 & 0.363 & 0.986 & 0.947 & 0.101 & \textbf{8.00} \\
\textbf{Ours} & \textbf{0.0558} & \textbf{30.3} & \textbf{0.339} & \textbf{0.988} & \textbf{0.959} & \textbf{0.0901} & \textbf{8.00} \\
\hline
\end{tabular}
\subcaption*{\textbf{8spp}}

\bigskip
\begin{tabular}{l|ccccccc}
\hline
\textbf{Method} & \textbf{L1 $\downarrow$} & \textbf{PSNR $\uparrow$} & \textbf{LPIPS $\downarrow$} & \textbf{DINO $\uparrow$}~\cite{oquab2023dinov2} & \textbf{CLIP $\uparrow$}~\cite{radford2021learningtransferablevisualmodels} & \textbf{FliP $\downarrow$}~\cite{Andersson2020} & \textbf{FoVVDP $\uparrow$}~\cite{10.1145/3450626.3459831} \\
\hline
AFGSA~\cite{Yu2021} & 0.0616 & 31.7 & 0.310 & 0.988 & 0.970 & 0.0760 & 8.42 \\
OIDN~\cite{OpenImageDenoise} & 0.0387 & \textbf{33.4} & 0.318 & 0.994 & 0.965 & 0.0634 & \textbf{8.93} \\
Ours & \textbf{0.0371} & \textbf{33.4} & \textbf{0.296} & \textbf{0.995} & \textbf{0.973} & \textbf{0.0589} & 8.89 \\
\hline
\end{tabular}
\subcaption*{\textbf{32spp}}

\label{tab:noisebase_metrics}
\end{table*}

\begin{figure*}
\centering
\includegraphics[width=\linewidth]{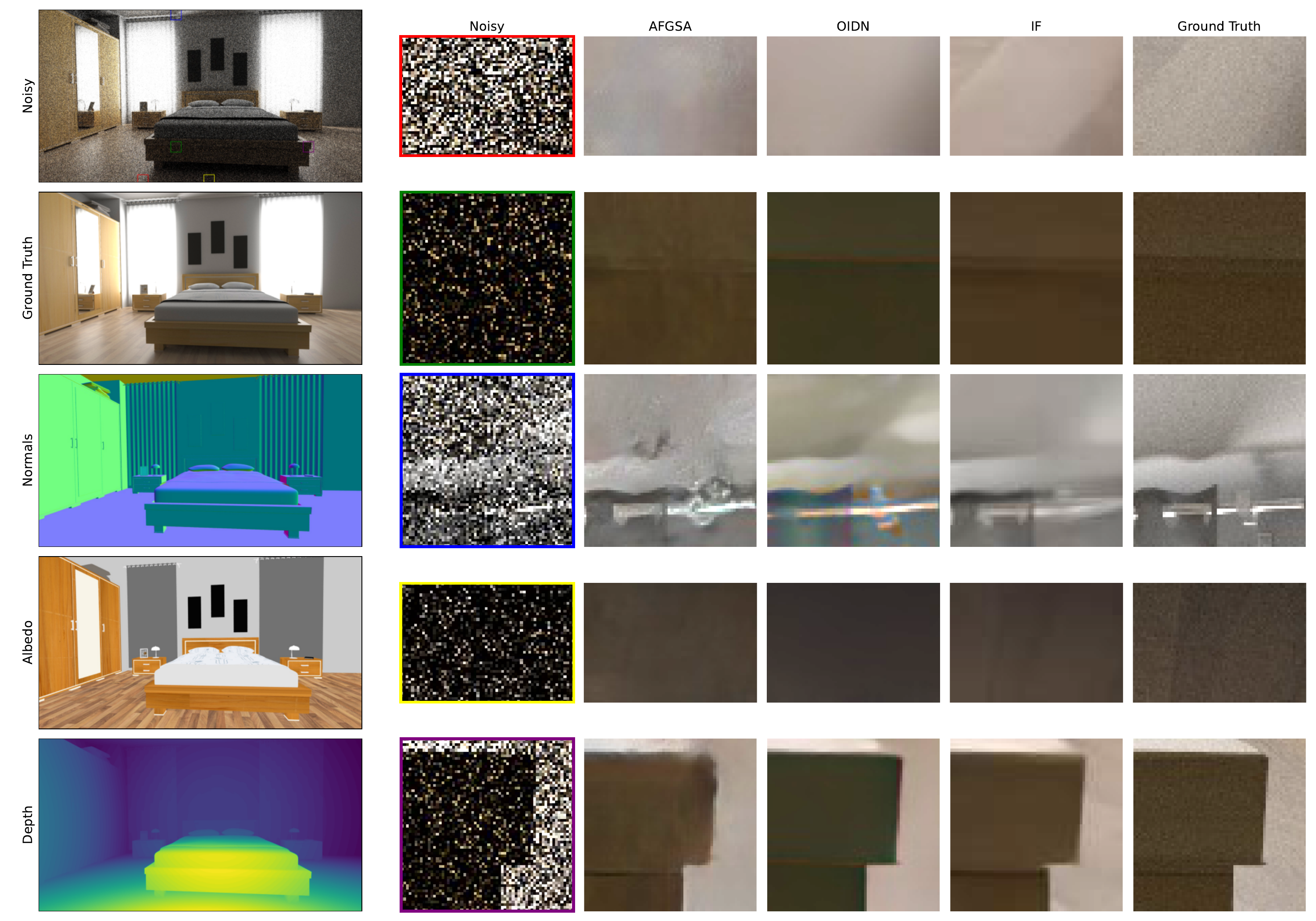}
\label{fig:sup8_1}
\caption{Additional qualitative results on the noisebase dataset. The first column shows the noisy radiance at 8 spp, reference, and auxiliary buffers. The noisy radiance has color-coded boxes of interest. \textbf{Columns 3 and 4} show results for competing methods. Observe how AFGSA generates noise because it uses an adversarial loss and the Ground Truth is a bit noisy. OIDN occasionally fails to reproduce color correctly. Our method (fifth column, \textbf{IF}) consistently produces reasonable results especially in undersampled regions.}
\end{figure*}

\begin{figure*}
\centering
\includegraphics[width=\linewidth]{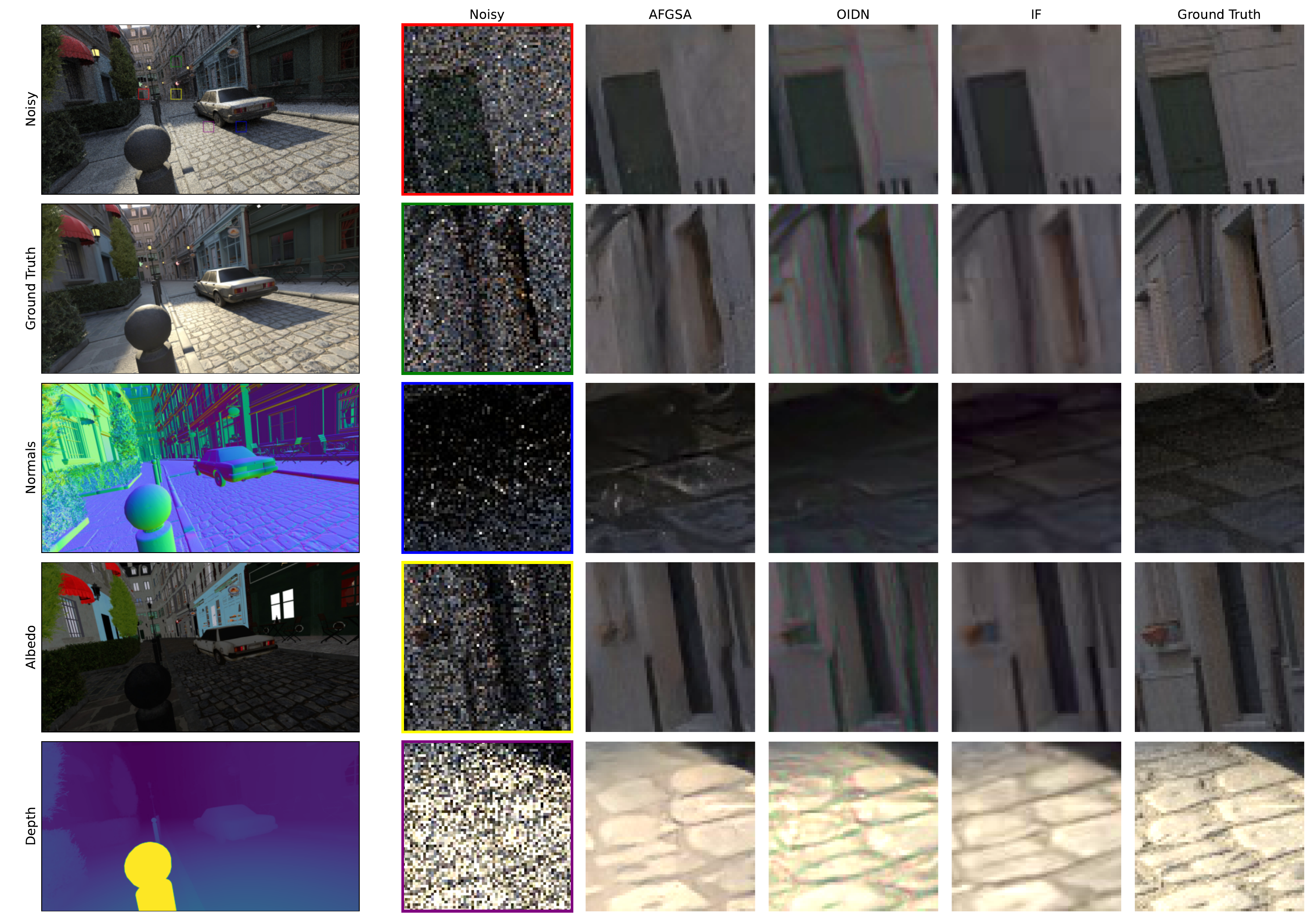}
\label{fig:sup8_2}
\caption{Additional qualitative results at 8 spp. }
\end{figure*}

\begin{figure*}
\centering
\includegraphics[width=\linewidth]{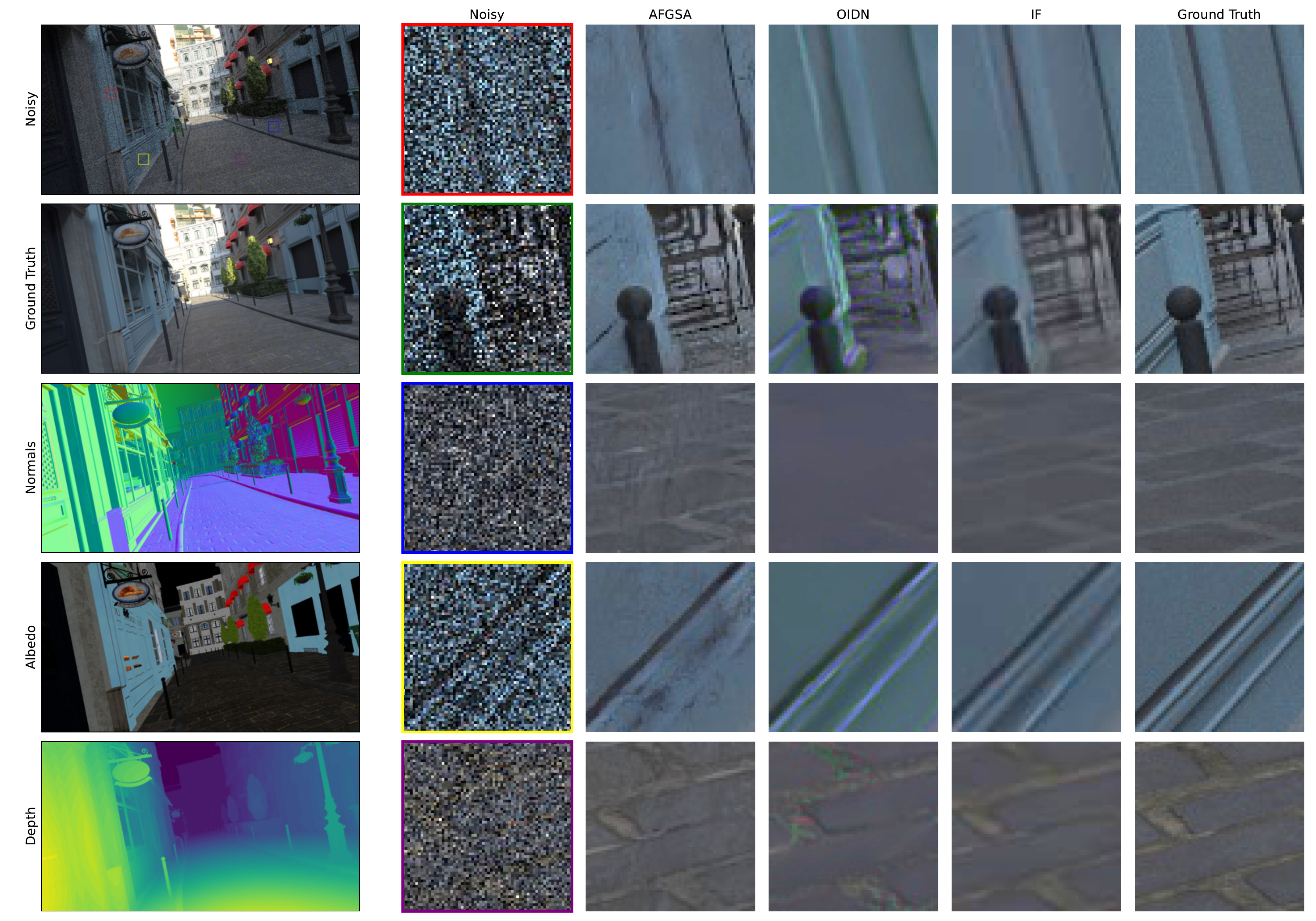}
\label{fig:sup8_3}
\caption{Additional qualitative results at 8 spp. }
\end{figure*}

\begin{figure*}
\centering
\includegraphics[width=\linewidth]{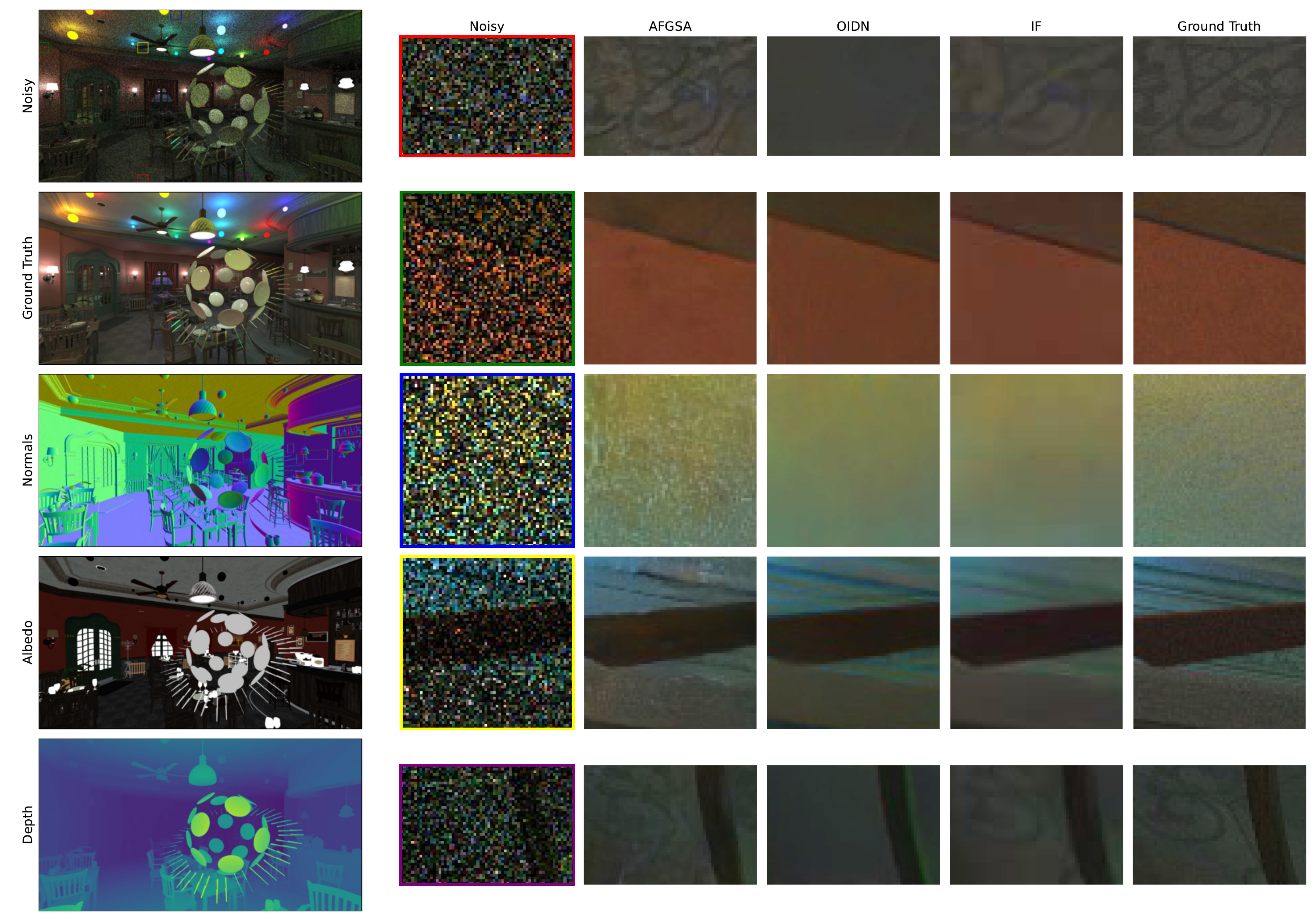}
\label{fig:sup8_4}
\caption{Additional qualitative results at 8 spp. }
\end{figure*}


\begin{figure*}
\centering
\includegraphics[width=\linewidth]{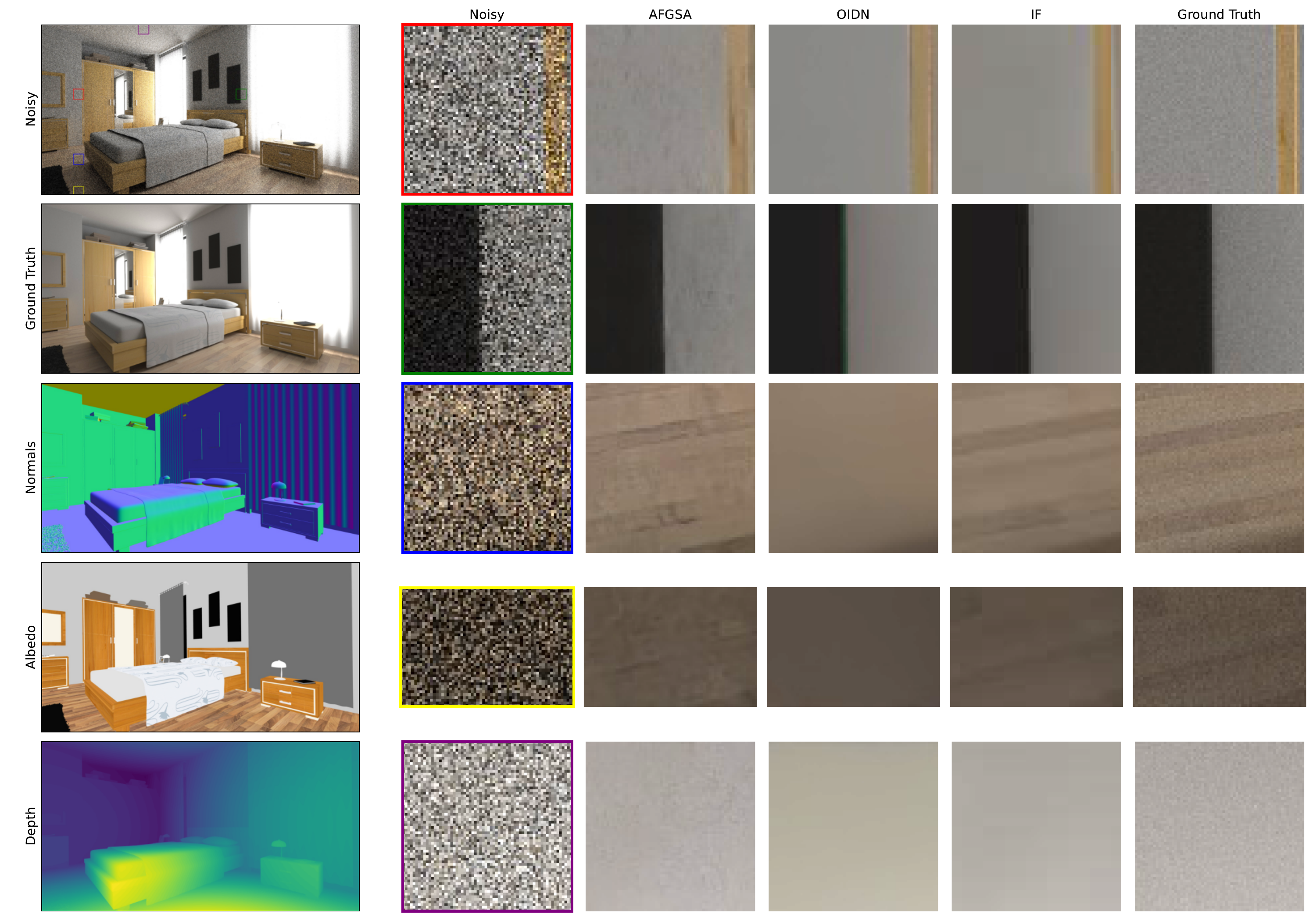}
\label{fig:sup32_1}
\caption{Additional qualitative results at 32 spp. }
\end{figure*}

\begin{figure*}
\centering
\includegraphics[width=\linewidth]{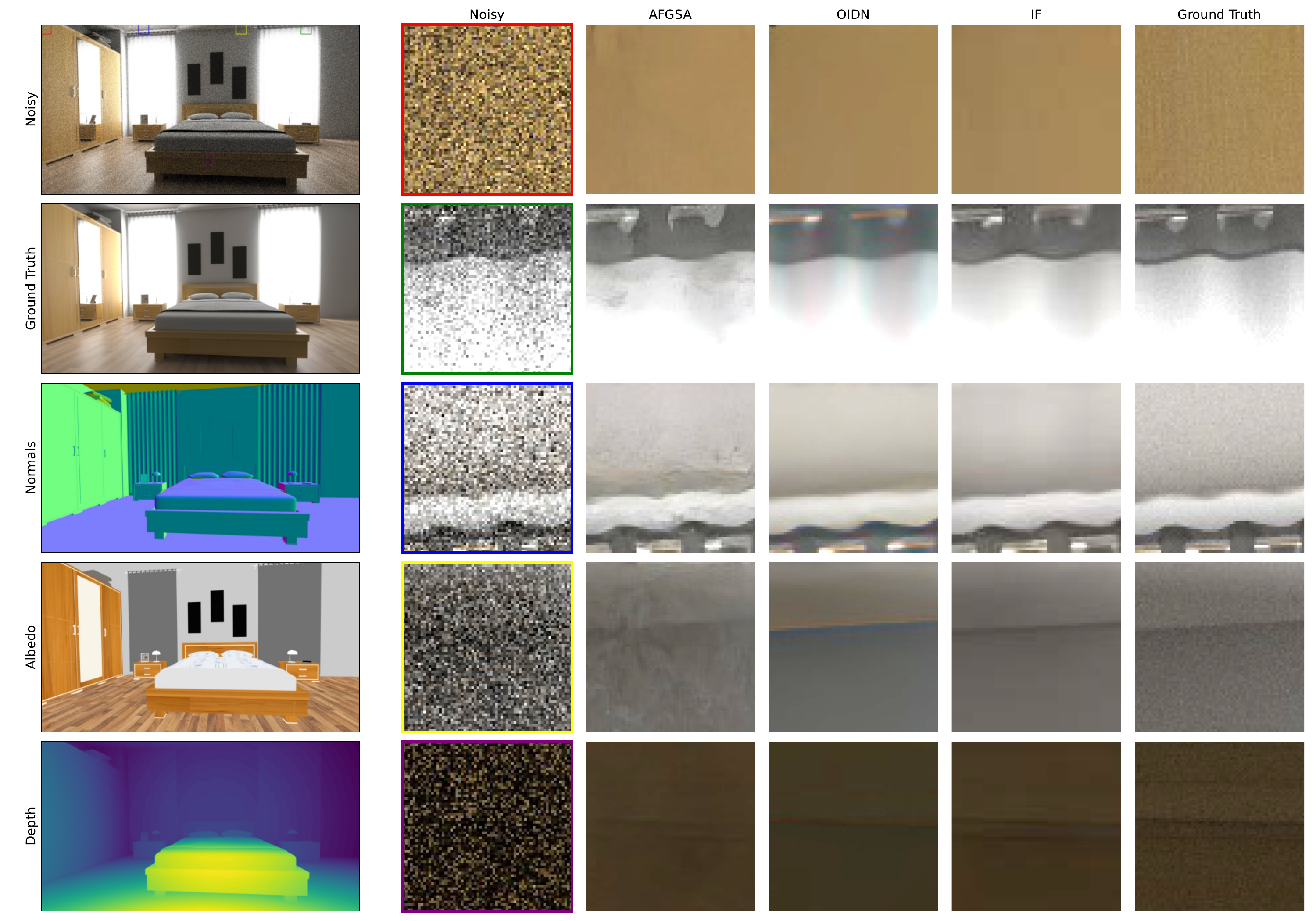}
\label{fig:sup32_2}
\caption{Additional qualitative results at 32 spp. }
\end{figure*}

\begin{figure*}
\centering
\includegraphics[width=\linewidth]{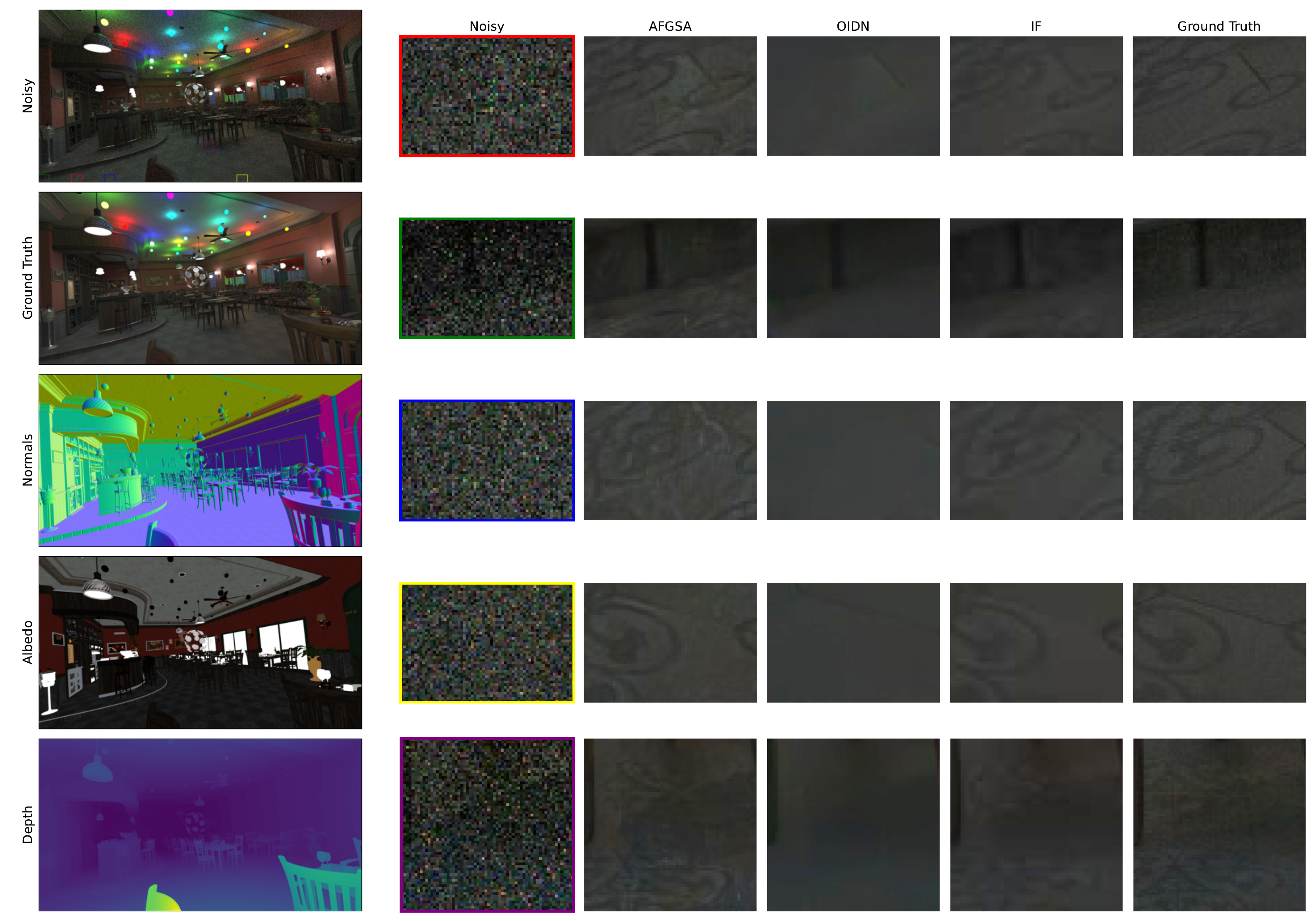}
\label{fig:sup32_3}
\caption{Additional qualitative results at 32 spp. }
\end{figure*}


\begin{figure*}
\centering
\includegraphics[width=\linewidth]{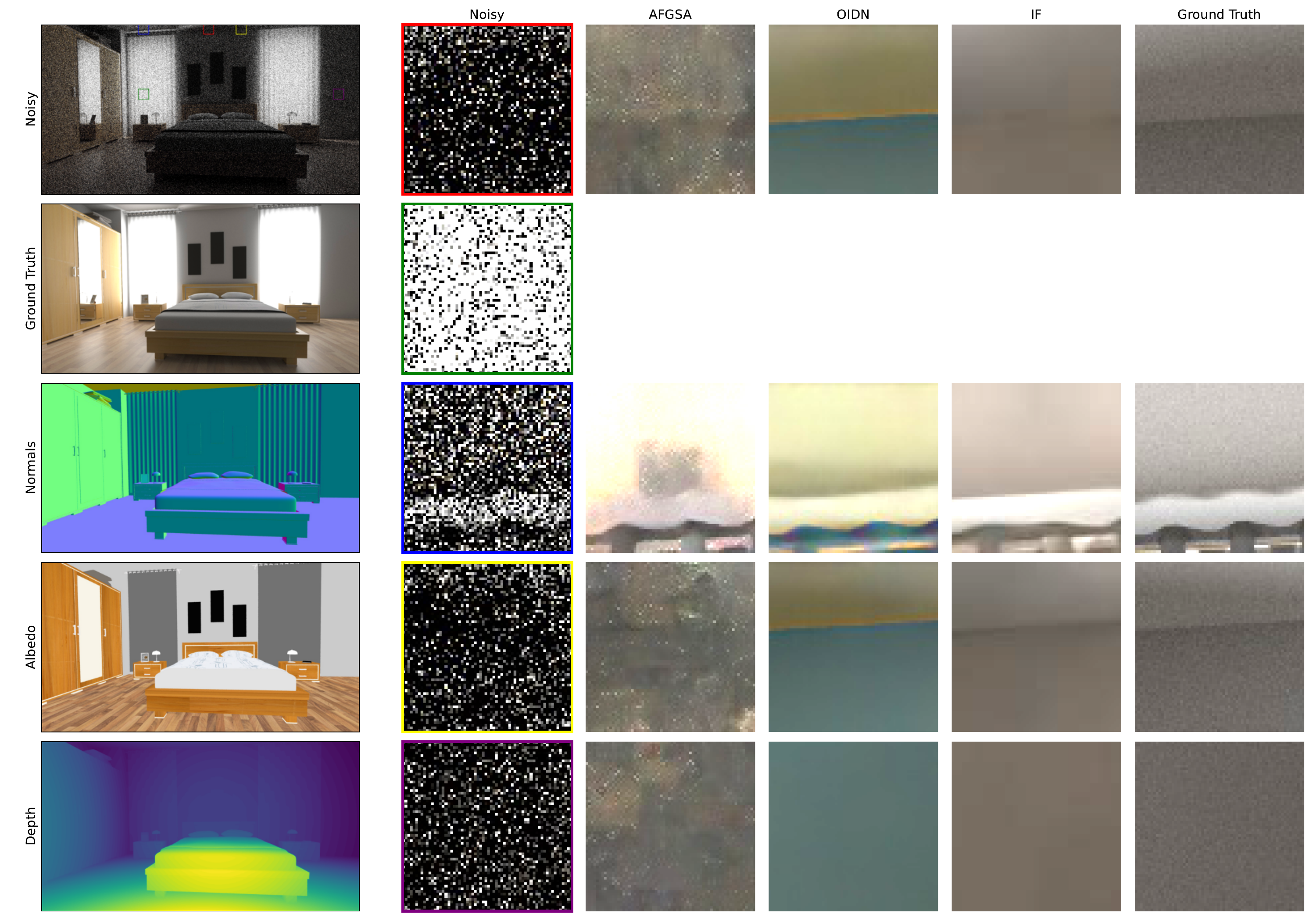}
\label{fig:sup2_1}
\caption{Additional qualitative results at 2 spp. }
\end{figure*}

\begin{figure*}
\centering
\includegraphics[width=\linewidth]{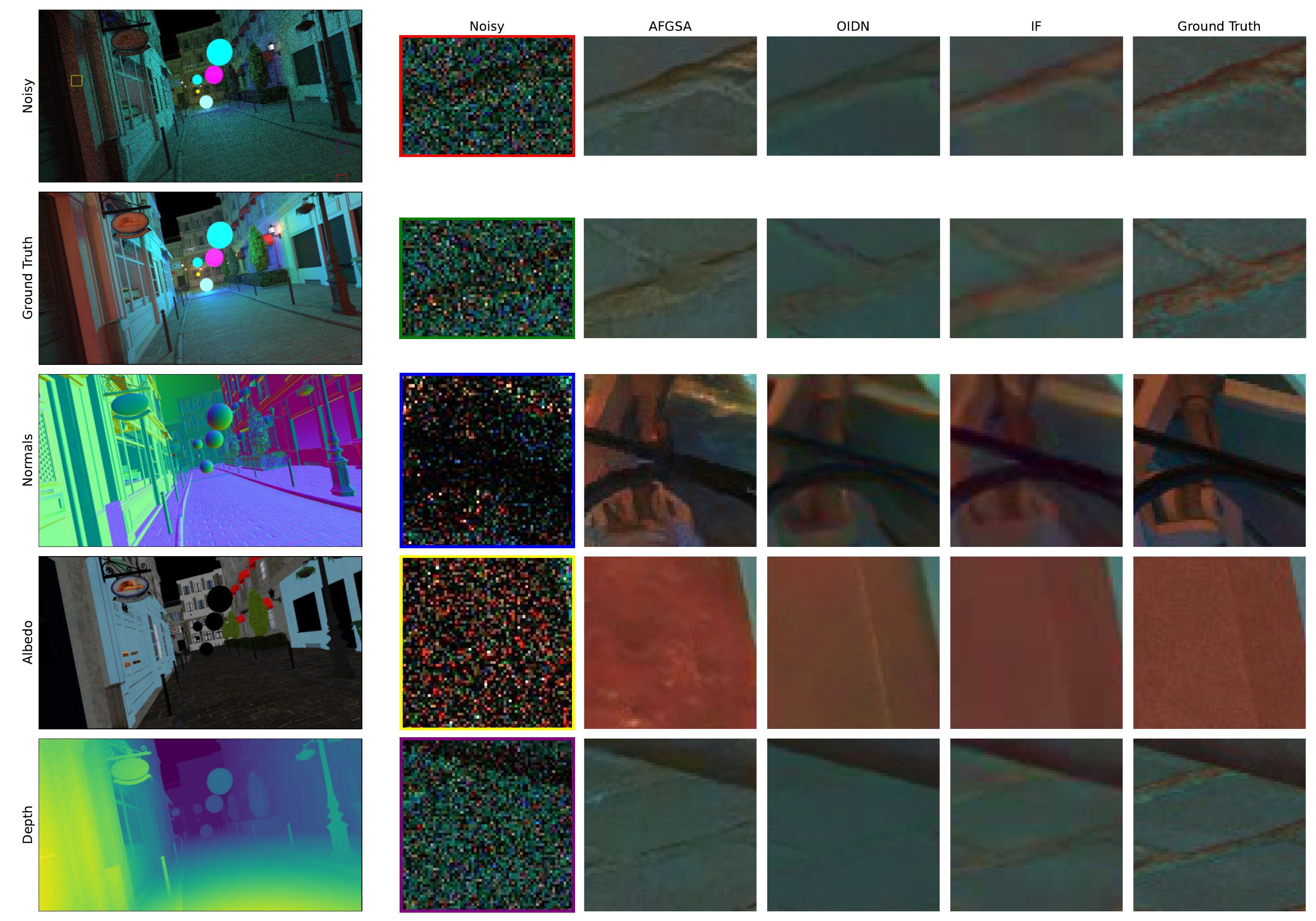}
\label{fig:sup2_2}
\caption{Additional qualitative results at 2 spp. }
\end{figure*}

\begin{figure*}
\centering
\includegraphics[width=\linewidth]{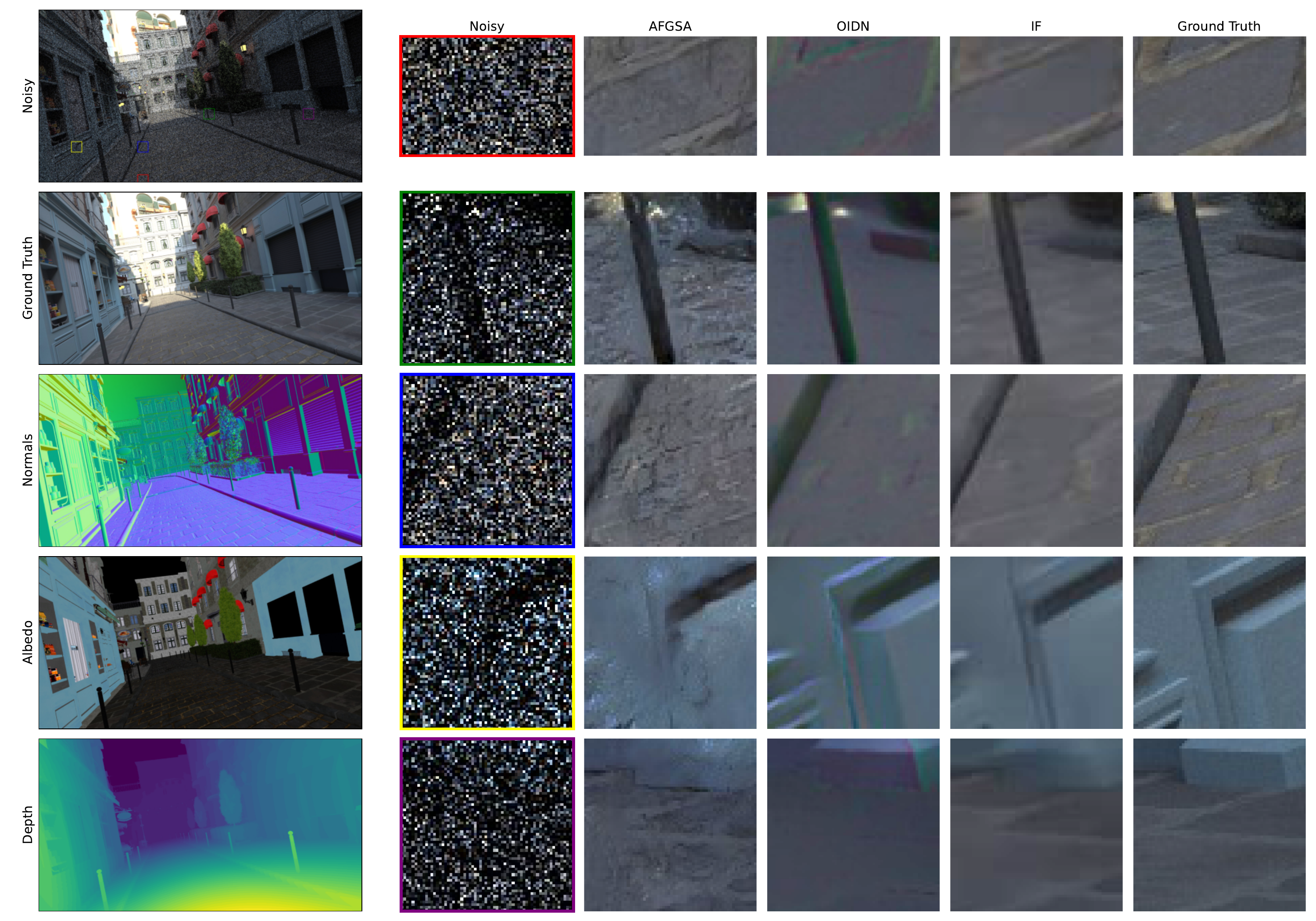}
\label{fig:sup2_3}
\caption{Additional qualitative results at 2 spp. }
\end{figure*}


\begin{figure*}
\centering
\includegraphics[width=\linewidth]{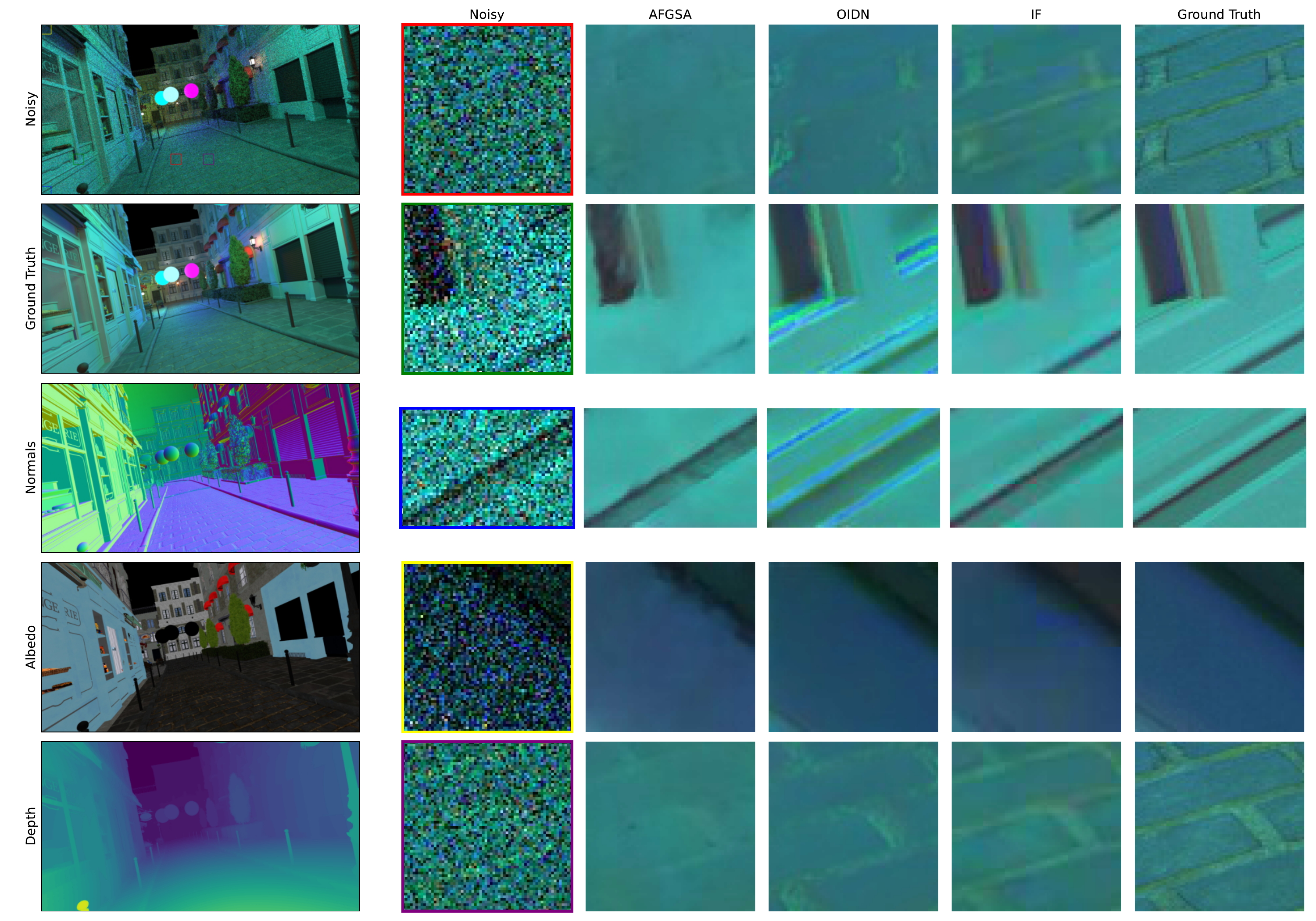}
\label{fig:sup4_1}
\caption{Additional qualitative results at 4 spp. }
\end{figure*}

\begin{figure*}
\centering
\includegraphics[width=\linewidth]{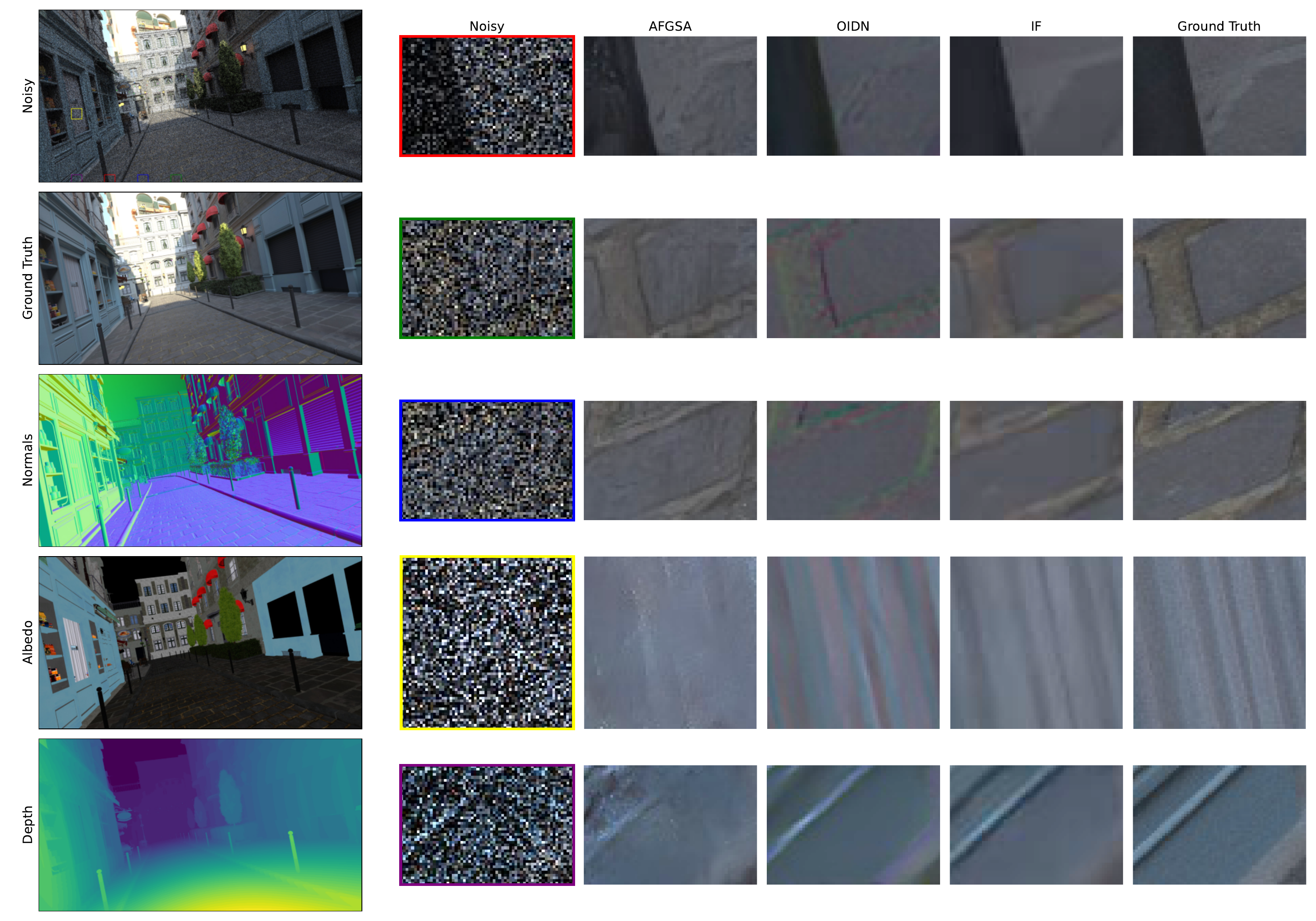}
\label{fig:sup4_2}
\caption{Additional qualitative results at 4 spp. }
\end{figure*}

\begin{figure*}
\centering
\includegraphics[width=\linewidth]{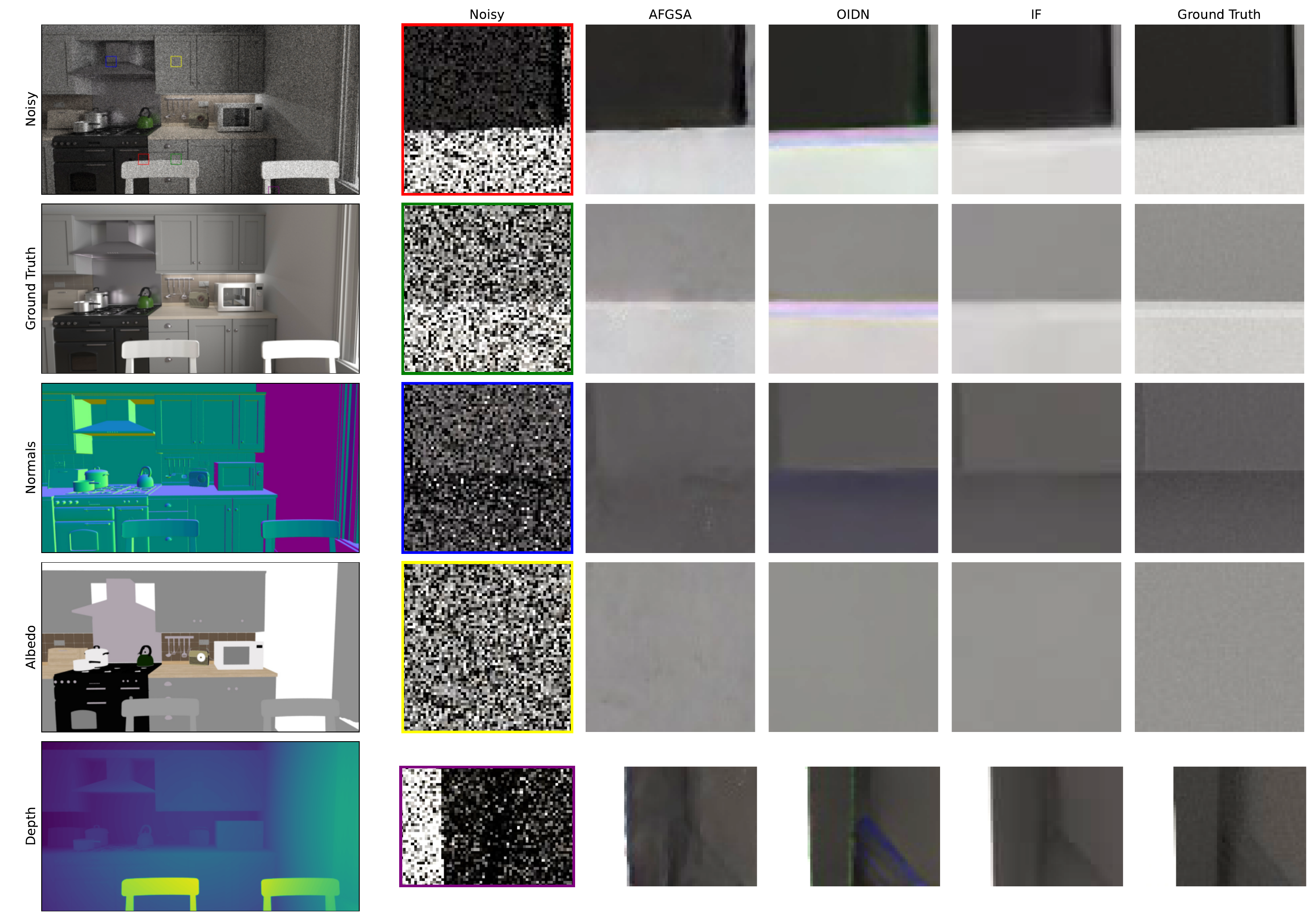}
\label{fig:sup4_3}
\caption{Additional qualitative results at 4 spp. }
\end{figure*}

\end{document}